\documentclass[twocolumn]{entalpic}

\usepackage{amsmath}
\usepackage{amssymb}
\usepackage{mathtools}
\usepackage[numbers,sort&compress]{natbib}
\usepackage{array}
\usepackage{url}

\newcommand{\atp}{\texttt{.atp}}
\newcommand{\code}[1]{\texttt{#1}}

\title{Atompack: A Storage and Distribution Layer for Read-Heavy Atomistic ML Training Datasets}

\author[1,2,*]{Ali Ramlaoui}
\author[1]{Daniel T. Speckhard}
\author[1]{Sagar Pal}
\author[2]{Fragkiskos D. Malliaros}
\author[1]{Alexandre Duval}
\author[1]{Victor Schmidt}

\affiliation[1]{Entalpic, Paris, France}
\affiliation[2]{Université Paris-Saclay, CentraleSupélec, Inria, Gif-sur-Yvette, France}
\contribution[*]{Corresponding author}
\metadata[Correspondence]{\email{\{ali.ramlaoui, daniel.speckhard, victor.schmidt\}@entalpic.ai}}

\abstract{
Atomistic machine learning datasets are increasingly used for training: large immutable snapshots are read repeatedly, shuffled across epochs, staged across clusters' storage systems, and republished as reusable scientific artifacts. This workload differs from interactive scientific curation, where mutable records and ad hoc inspection are often more important than random indexed throughput. We present Atompack, an append-oriented storage format and distribution layer designed around a simple workload: training pipelines usually consume complete molecular records, while the order of records is randomized by the learning algorithm. Atompack appends records efficiently during dataset construction, then commits an immutable index and serves records through a memory-mapped read path optimized for training. We compare Atompack with HDF5, LMDB, and ASE baselines representing array stores, key-value records, serialized records, and object-oriented databases. The benchmarks measure sequential reads, shuffled reads, shared-filesystem behavior, write throughput, and artifact size. On a representative 64-atom workload, Atompack is 96x faster than ASE LMDB on shuffled training-style reads while producing artifacts about 79\% smaller. The results indicate that serving complete molecule records, rather than field chunks or reconstructed objects, improves shuffled training throughput while keeping artifacts compact enough for public distribution.
}

\begin{document}

\maketitle

\section{Introduction}

Atomistic machine learning (ML) datasets are no longer only small collections of structures loaded for interactive analysis or small-scale training. Modern materials and molecular ML pipelines use millions to hundreds of millions of structures to train foundation interatomic potentials and related models~\cite{jain2013commentary, calderon2015aflow, kirklin2015open, draxl2019nomad, schmidt2024improving, barrosoluque2024openmaterials2024omat24, levine2025open, speckhard2025big, duval2024hitchhikersguidegeometricgnns, ramlaoui2025lemattrajscalableunifieddataset}. In that setting, the storage layer becomes part of the training system: records are repeatedly sampled in shuffled order, opened by multiple worker processes, and transferred between local and shared filesystems in high performance computing (HPC) environments. The same format also affects distribution: a published dataset should be easy for downstream users to download, reopen, and train from without reconstructing a project-specific database. The access pattern is structured but not always reflected in storage choices. A typical training objective has the form:
$$\min_{\theta} J(\theta) = \min_{\theta} \frac{1}{N}\sum_{i=1}^{N} \ell(f_{\theta}(x_i), y_i),$$
\noindent where \(\ell\) is a per-sample loss, \(x_i\) is a molecular system or crystal structure input that contains information about the elements present, the geometry, and other useful physical descriptors. \(x_i\) is typically encoded in a graph-based structure. Similarly, $y_i$ represents one or more targets such as energy, forces, stress, or charges. Stochastic training evaluates this sum through minibatches of graphs, using a static or dynamic batching algorithm, where the graphs are sampled with a permutation or near-permutation of the ids, \(\{1,\ldots,N\}\)~\cite{batching2026training}. Thus, the storage system is used to retrieve many fields describing the molecule/crystal \(\pi(i)\), not for one column over all molecules/crystals. This distinction matters because array stores perform best when nearby accesses reuse the same fields or chunks.

This workload differs in scope with respect to several common storage tools. The Atomic Simulation Environment (ASE) is a central interchange layer for atomistic workflows~\cite{ase-paper}, designed for object-level manipulation rather than high-throughput indexed reads. HDF5 provides compact scientific arrays and a mature ecosystem~\cite{hdf5}, strongest when access exploits array and chunk locality. LMDB provides an embedded key-value store with memory-mapped access and is widely used in training pipelines~\cite{chu2011mdb}, but the record payload format is left to each pipeline. Atompack targets a more specific problem: serving immutable atomistic datasets efficiently during training.

Atompack stores the unit consumed by training: a complete molecule or crystal structure. Its lifecycle is append during construction, commit an immutable index, distribute, and reopen read-only for training. Atompack therefore uses append-oriented construction, a committed index for \(O(1)\) record lookup, memory-mapped read-only serving, and repository-friendly shard directories. These choices are conservative: they trade away arbitrary updates and field projection to optimize the access pattern that dominates large-scale training.

We make three contributions: an Atompack file/shard design for immutable atomistic records; a comparison with HDF5, LMDB, and ASE-backed baselines across read, write, storage-size, and filesystem-sensitive workloads; and an open implementation with benchmark scripts and packaged datasets.

\begin{figure*}[t]
  \centering
  \includegraphics[width=0.97\textwidth]{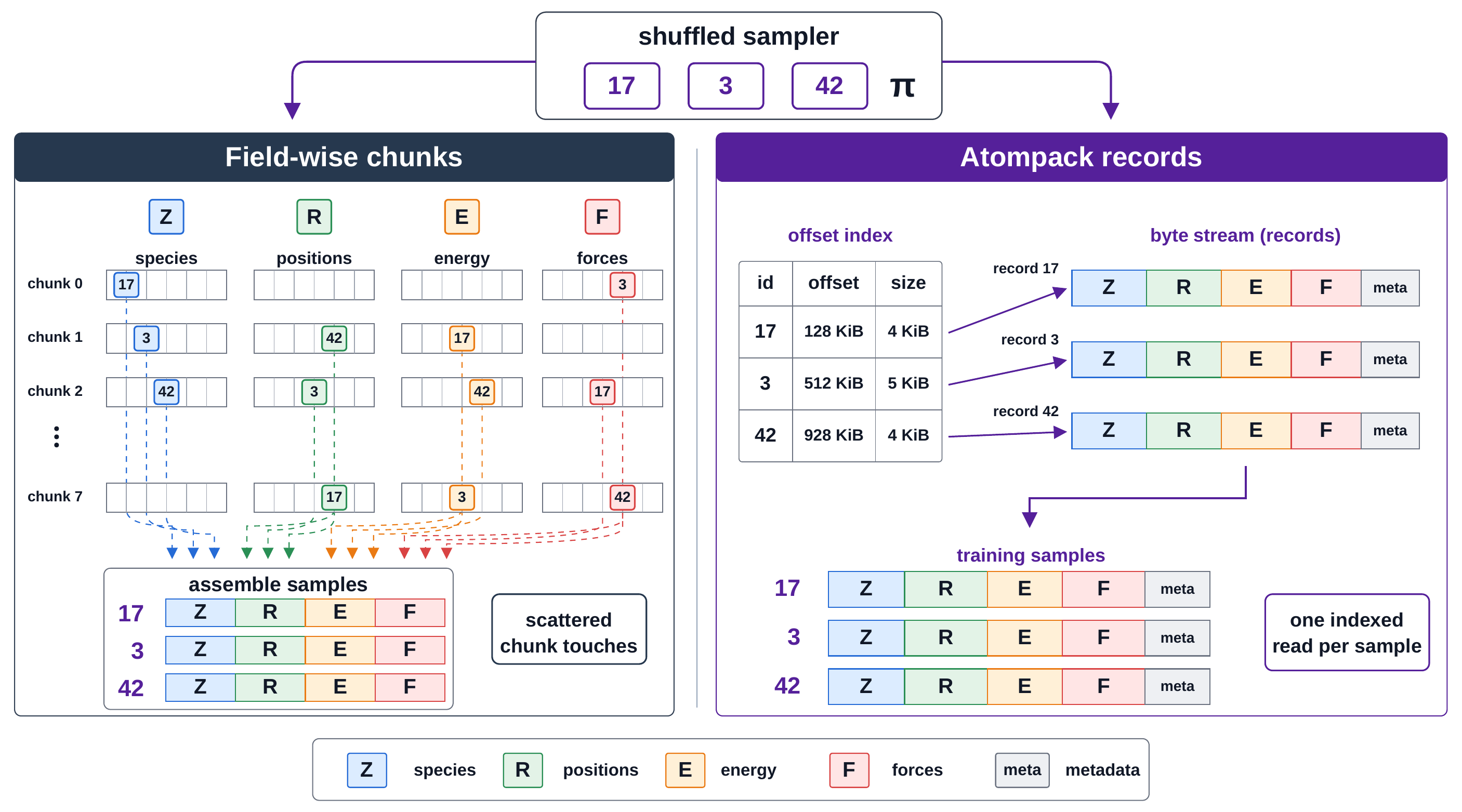}
  \caption{Storage access pattern for shuffled atomistic ML training. The left panel illustrates a field-wise chunked layout like HDF5: species, positions, energies, and other fields are stored as separate datasets, and the highlighted cells mark the same shuffled samples across those datasets. Atompack instead stores each molecule or structure as one indexed record, so the same shuffled sampler performs one direct record lookup per sample and returns the complete training input.}
  \label{fig:architecture}
\end{figure*}

\section{Background and Related Work}

\subsection{Atomistic Data Infrastructure}

Atomistic simulation software has long emphasized interoperability and rich structure objects. ASE is a representative example: it provides a high-level interface for building, manipulating, simulating, and storing atomic structures~\cite{ase-paper}. This object model is useful for curation and simulation integration. In stochastic ML training, however, the model consumes typed tensors, and repeated object materialization can cost as much as the storage access. Atompack keeps ASE-compatible ingestion paths, but serves training data as indexed tensor payloads.

Large public atomistic datasets have also changed expectations. The Open Catalyst 2020 dataset and related challenges demonstrated the value of large-scale open simulation data for ML-driven chemistry~\cite{Chanussot_2021}. Materials datasets such as MPtrj~\cite{deng2023chgnetpretraineduniversalneural} and OMat24~\cite{barrosoluque2024openmaterials2024omat24} further illustrate the growth of reusable potential-energy-surface data for training universal ML interatomic potentials~\cite{siron2025lematbulkaggregatingdeduplicatingquantum,kaplan2025matpes,kuner2025mpaloe}. These datasets are valuable only if researchers can fetch, reopen, and iterate over them efficiently. For these datasets, the published storage layout affects reproducibility and downstream reuse.

\subsection{Scientific Storage and Training Stores}

HDF5 is a standard choice for scientific arrays and is well suited to structured, typed datasets with chunked access~\cite{hdf5}. Zarr and related formats extend this array-store lineage toward object and cloud storage~\cite{zarrspec}. These formats are strong choices when the workload can exploit field-wise or chunk-wise locality. Atomistic ML training often asks a different question: retrieve all fields for molecule \(i\), then molecule \(j\), under a shuffled index stream. A structure-of-arrays layout minimizes storage overhead and enables efficient scans, but a shuffled complete-record workload can turn each sample into several independent chunk lookups and reconstruction steps.

LMDB is widely used in ML pipelines because it provides direct key-value lookup through an embedded memory-mapped store~\cite{chu2011mdb}. In practice, however, the record payload is project-specific. A packed binary payload can be compact but requires custom schema code; a pickled dictionary is easier to build but pays serialization and object reconstruction costs. General ML frameworks such as PyTorch provide data-loading abstractions~\cite{paszke2019pytorch}, but they intentionally leave storage layout to the dataset author. Atompack instead uses one structure per indexed record and a domain-specific schema for atomistic descriptors/targets.

\section{System Design}

Figure~\ref{fig:architecture} shows the intended lifecycle. Atompack targets the transition from dataset construction to repeated training access. Before publication, data are generated and appended; after publication, the dominant operation is read-only indexed access by training workers.

\begin{figure*}[t]
  \centering
  \includegraphics[width=\linewidth]{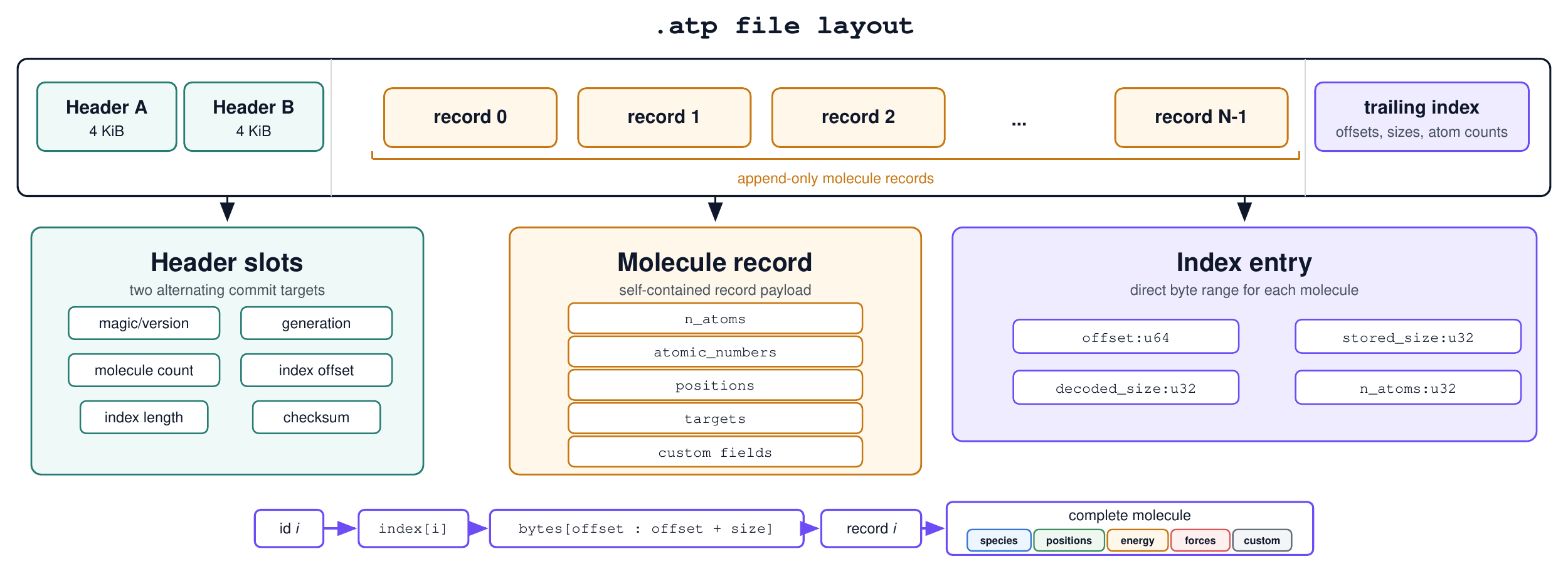}
  \caption{The \atp{} layout separates append-friendly record writes from indexed reads. A committed trailing index maps molecule indices to byte offsets and record sizes, enabling direct lookup in read-only mode. The figure illustrates the layout for molecular records; crystal-structure records use the same layout with additional fields for the cell and periodic boundary conditions.}
  \label{fig:layout}
\end{figure*}

\subsection{Workload Model and Design Rationale}

Let a dataset contain \(N\) molecules. Molecule \(i\) has a byte payload of size \(s_i\), containing geometry and targets, and training reads an index sequence \(\pi\).
For an indexed record layout, one epoch performs \(N\) record lookups. Each lookup uses one \(O(1)\) offset-table probe and reads \(s_{\pi(i)}\) record bytes, so the ideal storage work for one epoch is:
\[
  N O(1) + O\!\left(\sum_{i=1}^{N} s_{\pi(i)}\right)
    =
    N O(1) + O\!\left(\sum_{i=1}^{N} s_i\right)
\]
because \(\pi\) is a permutation.
The total record bytes read are therefore independent of sample order. The design question Atompack addresses is how much indexing, reconstruction, and locality overhead the layout adds when that order is shuffled.
Storing all input and target fields for a sample in one record aligns I/O with minibatch sampling.

A chunked field-wise layout has a different cost: suppose a field is divided into \(C\) balanced chunks and a minibatch contains \(b\) randomly selected molecules. The relevant quantity is not only the number of molecules in the batch, but how many chunks must be touched to assemble that batch.

For a fixed chunk \(c\), the probability that none of the \(b\) sampled molecules falls in that chunk is \((1-1/C)^b\). Thus the probability that chunk \(c\) is touched is \(1-(1-1/C)^b\), and by linearity of expectation over the \(C\) chunks, the expected number of touched chunks is
\[
  C\left(1-\left(1-\frac{1}{C}\right)^b\right).
\]
When \(b \ll C\), collisions between samples are rare, so this expectation is close to \(b\). With several fields, random complete-record sampling can therefore lose the amortization that makes chunked arrays attractive for scans. This behavior reflects the workload rather than a limitation of HDF5 or Zarr: shuffled samples do not reuse field chunks.

This model motivates four design choices. First, Atompack uses molecule-granularity records rather than arbitrary row groups or columns. Second, records are indexed by a compact offset table, so lookup is independent of dataset size once the index is available. Third, construction is append-oriented and read serving is immutable, matching the common build/freeze/train lifecycle of ML corpora~\cite{aizman2020highperformanceiolarge}. Fourth, the same artifact can be a local file, a shard directory, or a repository path, so distribution does not require downstream users to reconstruct the storage layout.

\begin{table*}[ht]
\caption{Design Choices and Consequences.}
\label{tab:design}
\centering
\scriptsize
\begin{tabular}{p{0.22\textwidth}p{0.40\textwidth}p{0.30\textwidth}}
\toprule
Choice & Rationale & Cost \\
\midrule
Whole-molecule records & Aligns storage unit with the loss sample \(x_i\) & Less efficient for field-only analytics \\
\addlinespace[0.8em]
Trailing offset index & \(O(1)\) direct lookup after opening & Requires committed immutable snapshots \\
\addlinespace[0.8em]
Append/freeze lifecycle & Matches build, publish, train workflows & Updates/deletes require rewriting \\
\addlinespace[0.8em]
Memory-mapped read mode & Shares immutable bytes across workers and avoids global rescans & Performance depends on filesystem paging behavior \\
\addlinespace[0.8em]
Shard directories & Supports repository-scale datasets and partial distribution & Cross-shard routing must be explicit \\
\bottomrule
\end{tabular}
\end{table*}

\subsection{Schema and Distribution Model}

The schema is domain-specific but not model-specific. Required fields represent molecular geometry; optional fields cover common potential-energy-surface targets such as energies, forces, stress, cells, charges, velocities, and periodic boundary conditions. Custom per-molecule and per-atom properties allow a dataset to add targets without changing the core format. The schema is narrower than a general scientific database but less project-specific than arbitrary binary records. A fully general scientific database would permit richer mutation and queries but impose machinery that is not needed during training. A collection of project-specific binary records can be fast but makes reuse, validation, and public distribution harder. Atompack uses a small common schema so data producers and consumers can agree on the representation of common atomistic ML quantities while preserving room for dataset-specific properties.

Distribution is part of the storage design. Atompack supports a single \atp{} file for local artifacts and shard directories for larger repositories. A public repository can expose a flat logical dataset even when the underlying object is sharded.

\subsection{File Layout and Operational Modes}

The \atp{} file layout, shown in Figure~\ref{fig:layout}, consists of two fixed-size header slots, an append-only data region, and a trailing index. Each header stores the format version, generation number, index offset and length, molecule count, record and codec metadata, and a checksum. The two header slots form a small commit protocol: when committing a new index, the writer updates the inactive header slot, leaving the previous valid header available if the write is interrupted. On open, Atompack validates both headers and selects the newest valid generation. Records are appended sequentially, a new index is written, and the writer then advances the valid header generation.

Each molecule is encoded as a typed record payload containing its required geometry fields, optional targets, and custom properties. The payload is then stored directly or compressed with the selected per-file codec. The resulting record bytes are concatenated in the append-only data region, while the trailing index stores each record's byte offset, stored size, uncompressed size, and atom count. Reads therefore use the index to select a byte range, decompress it if needed, and decode only the requested molecule.

Two open modes are deliberately distinct. Writable mode appends records and commits new trailing indexes. Read-only mode uses memory mapping for static datasets. This mode is the intended serving path for training: workers can reopen the same immutable snapshot, fetch structures by index, and avoid rescanning or reparsing global metadata for each access.

The format is specialized. It stores whole molecules or structures rather than arbitrary field projections, and updates or deletes require rewriting a dataset. That trade-off is intentional: atomistic ML training normally consumes complete molecule records, and immutable snapshots are easier to publish, cache, and benchmark. The write-throughput results in Section~\ref{sec:results} suggest that rewriting is practical at the measured write rates.

\section{Evaluation Methodology}
\label{sec:eval}

We evaluate Atompack as a practical data-serving layer rather than as a microbenchmark of compression or serialization alone. The comparison is organized around storage methods. HDF5 SOA represents compact field-wise scientific arrays, chunked along the molecule dimension with chunks of 256 molecules. LMDB Packed is a purpose-built optimized indexed key-value design with a compact per-molecule binary payload. LMDB Pickle represents a common convenience choice: key-value lookup plus Python-level serialization. ASE SQLite and ASE LMDB represent object-oriented atomistic database access, with ASE LMDB serving as the closest community-standard reference point. The baselines represent four storage choices used in practice: array stores, indexed key-value records, Python-serialized records, and rich atomistic object stores. Unless otherwise noted, the reported read-throughput, write-throughput, and
storage-footprint results use uncompressed Atompack and LMDB payloads
(\code{codec=none}). They also reflect common public-dataset workflows: OC20 uses LMDB-backed training datasets in the Open Catalyst ecosystem~\cite{Chanussot_2021}, OMat24 is a recent large-scale FAIR Chemistry materials corpus distributed for atomistic ML reuse~\cite{barrosoluque2024openmaterials2024omat24}, and OMol25 is provided as ASE DB files using the LMDB backend (\code{*.aselmdb})~\cite{levine2025open,fairchemomol25docs}.

\begin{table}[t]
\caption{Evaluation Matrix}
\label{tab:matrix}
\centering
\scriptsize
{\renewcommand{\arraystretch}{1.18}
\begin{tabular}{p{0.24\linewidth}p{0.31\linewidth}p{0.32\linewidth}}
\toprule
Workload & Configuration & Metric \\
\midrule
Sequential read & 12, 64, 256 atoms/molecule; controlled synthetic records & molecules/s, atoms/s, 95\% CI \\
\addlinespace[0.35em]
Random/shuffled read & single-worker multiprocessing slice; same random index list per backend & molecules/s, atoms/s, 95\% CI \\
\addlinespace[0.35em]
Filesystem sensitivity & 64 atoms/molecule on NVMe, NFS, GPFS, Lustre & atoms/s and backend speedup \\
\addlinespace[0.35em]
Write throughput & built-in fields and synthetic custom-property rows & molecules/s, 95\% CI \\
\addlinespace[0.35em]
Storage footprint & artifacts produced by write benchmark & size ratio vs Atompack \\
\bottomrule
\end{tabular}}
\end{table}

The controlled read datasets contain fixed-size molecules with representative atomistic fields. The main atom counts are 12, 64, and 256 atoms per molecule. High-throughput backends use datasets with up to \(10^6\) molecules; slow ASE-backed baselines use smaller generated datasets where needed to keep benchmark runtime tractable, while preserving the same per-sample payload and access semantics. Sequential reads use contiguous indices, while random/shuffled reads sample molecule indices without replacement using a fixed seed. Benchmark functions touch representative arrays and scalar fields so that results do not measure only dictionary lookup or index arithmetic. The multiprocessing read benchmark uses persistent workers. For the main random-read figures, we report the single-worker shuffled-read measurement, which best isolates the storage cost of training-style random access.

The write benchmark separately measures records with only common potential-energy-surface fields and records with additional custom properties. This distinction matters because some storage layouts handle fixed schemas and extension fields differently. Storage footprint is reported as normalized artifact size relative to Atompack for the same generated records. Write-throughput rows use one warmup trial followed by three timed trials. Single-process read rows use one warmup pass followed by five timed trials.
Worker-backed shuffled-read rows use one warmup pass followed by three timed
trials. The worker-backed benchmark uses fewer repetitions because it is evaluated over more backend and is more expensive. The plots report medians with 95\% confidence intervals where available in the result rows.

The storage environments represented in the maintained result artifacts are a Samsung 990 EVO Plus NVMe SSD, NFSv3 shared storage, GPFS, and Lustre filesystems. This range reflects common scientific ML deployment paths: datasets may be generated or converted on local solid-state storage, staged through laboratory shared storage, and trained from parallel filesystems on HPC systems.

Cold versus warm-cache state can change random-read throughput. Access path also matters: single-record reads, one-element batches, and native flat batches do not exercise identical overhead. Finally, materializing rich atomistic objects can become the bottleneck and reduce the visible difference between storage backends. Accordingly, each result specifies the read path, and we avoid comparing object-materialization timings directly with array-serving timings.

\section{Benchmark Results}
\label{sec:results}

\subsection{Read Throughput}

\begin{figure*}[t]
  \centering
  \includegraphics[width=0.94\textwidth]{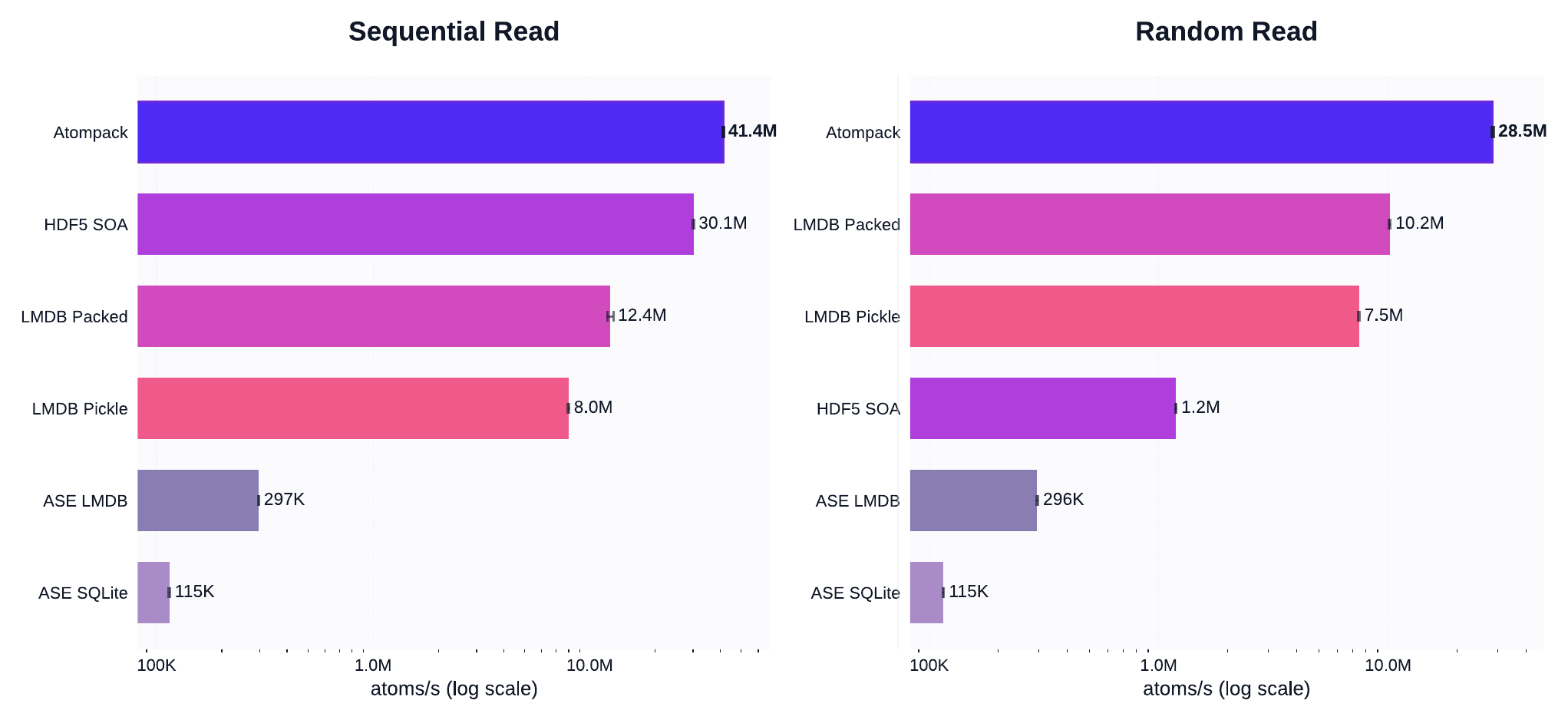}
  \caption{Read throughput for a dataset of 64-atom molecules randomly generated on an NVMe workload using a Samsung 990 EVO Plus SSD. The random-read panel uses the single-worker shuffled-read slice and shows the largest practical separation between storage layouts. Log-scale on the x-axis is used to visualize the order of magnitude differences in read speeds; on random reads, Atompack is 24.0\(\times\) faster than HDF5 SOA and 3.82\(\times\) faster than LMDB Pickle.}
  \label{fig:readhero}
\end{figure*}

Figure~\ref{fig:readhero} shows the main 64-atom NVMe read comparison. Atompack reaches 646,261 molecules/s for sequential reads and 445,830 molecules/s for the shuffled single-worker path. The sequential result is 1.37\(\times\) faster than HDF5 SOA, 3.32\(\times\) faster than LMDB Packed, and 5.18\(\times\) faster than LMDB Pickle for the same slice. The larger separation appears in the random/shuffled path: Atompack is 24.0\(\times\) faster than HDF5 SOA, 2.81\(\times\) faster than LMDB Packed, and 3.82\(\times\) faster than LMDB Pickle.

The HDF5 result illustrates the difference between compact array storage and shuffled molecule serving. HDF5 SOA remains strong for sequential access, but the random path loses chunk locality and requires per-sample reconstruction. LMDB layouts are closer to the molecule-indexed access pattern, but they still pay record-specific decode costs and language-level payload reconstruction. ASE-backed stores are slower mainly because of database lookup and row materialization: each access retrieves a full row, decodes array payloads and metadata into an \code{AtomsRow}, and only then exposes the fields used by the benchmark. This tradeoff prioritizes a rich scientific object interface over raw indexed throughput.

\subsection{Shared Filesystems}

\begin{figure}[t]
  \centering
  \includegraphics[width=\linewidth]{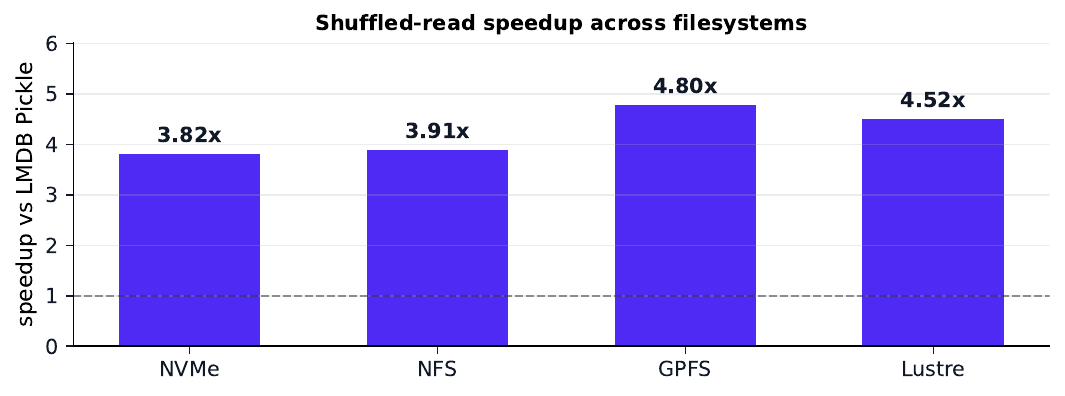}
  \caption{Random/shuffled read speedup over LMDB Pickle across filesystems for the 64-atom workload, computed as Atompack throughput divided by LMDB Pickle throughput on the same filesystem. Atompack maintains a 3.82--4.80\(\times\) advantage across local and shared storage, which is important for cluster-based training and dataset reuse.}
  \label{fig:filesystems}
\end{figure}

Most large-scale training pipelines do not train directly from a local NVMe disk. Shared filesystems introduce metadata and I/O behavior that can change the relative ranking of storage layouts. Figure~\ref{fig:filesystems} compares the 64-atom shuffled single-worker path across NVMe, NFS, GPFS, and Lustre by normalizing Atompack throughput against LMDB Pickle, a common serialized-object key-value layout. Atompack is 3.82\(\times\) faster on NVMe, 3.91\(\times\) on NFS, 4.80\(\times\) on GPFS, and 4.52\(\times\) on Lustre. In absolute terms, Atompack reaches 445,830 molecules/s on NVMe, 353,485 on NFS, 247,931 on GPFS, and 353,195 on Lustre.

The filesystem benchmark matches a common deployment path: local conversion, shared-storage staging, and HPC training. A storage format that only performs well on local disks may still create bottlenecks for realistic training infrastructure. Atompack's direct index and whole-molecule record layout give it consistent behavior under this range of deployment conditions.

\subsection{Write Throughput and Storage Footprint}

\begin{figure*}[t]
  \centering
  \includegraphics[width=0.92\textwidth]{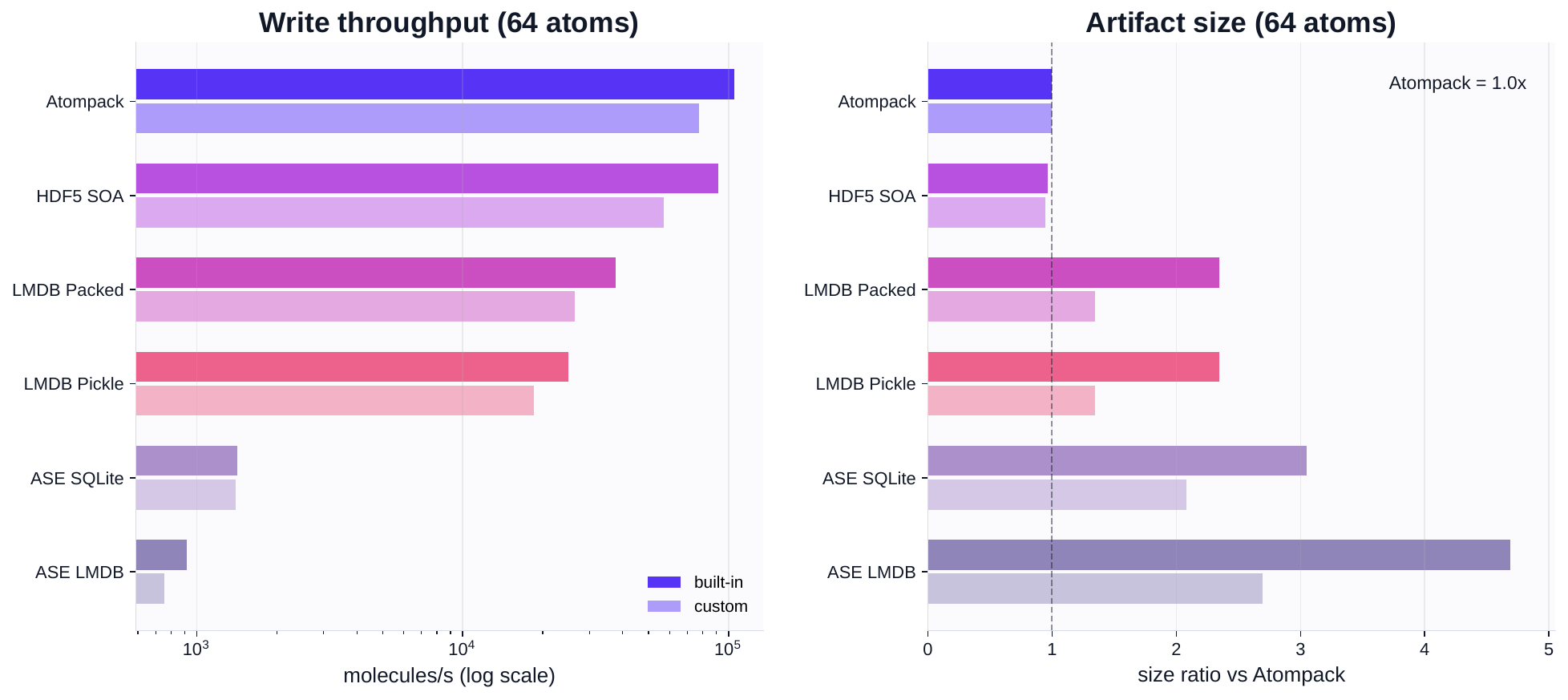}
  \caption{Write throughput and storage footprint for 64-atom records. The write-throughput axis is log scaled. Atompack combines the highest measured write throughput with storage size close to compact HDF5 SOA and smaller artifacts than LMDB and ASE-backed baselines in this benchmark.}
  \label{fig:writestorage}
\end{figure*}

Read throughput alone is not enough: large atomistic datasets are often regenerated, filtered, converted, and republished. Figure~\ref{fig:writestorage} summarizes write throughput and artifact size for the 64-atom NVMe write benchmark. Atompack writes 105,473 molecules/s for built-in fields and 77,193 molecules/s when synthetic custom properties are included. HDF5 SOA is competitive on built-ins at 91,431 molecules/s and 57,198 molecules/s with custom properties. LMDB Packed and LMDB Pickle are substantially slower in these write slices, and ASE-backed stores are slower by roughly two orders of magnitude.

Storage size shows the intended compromise. HDF5 SOA is slightly smaller than Atompack in the 64-atom slice, at 0.96\(\times\) Atompack size for built-ins and 0.95\(\times\) with custom properties. Atompack remains near that compact array-oriented regime while LMDB Packed and Pickle are 2.34--2.35\(\times\) Atompack size for built-ins and 1.35\(\times\) with custom properties. ASE SQLite and ASE LMDB are larger still in the normalized comparison. Atompack is slightly larger than HDF5 SOA in this slice, but has much higher shuffled-read throughput; compared with LMDB and ASE-backed stores, it also reduces artifact size.

The atom-count scaling results also suggest why larger systems are a distinct regime. Fixed per-record metadata and serialization overheads are amortized as molecule payloads grow, so small and medium records emphasize indexing and serialization overheads, while larger records increasingly expose payload movement and filesystem bandwidth. This trend helps explain why compact whole-record storage remains useful even when the relative storage-size gap to key-value layouts narrows for larger molecules.

The three result slices point to the same design tradeoff. HDF5 SOA is slightly more compact than Atompack but far lower on shuffled reads. LMDB Packed and LMDB Pickle improve random lookup relative to HDF5 SOA but require larger artifacts and lower write throughput in the representative slice. ASE-backed stores are useful for scientific object workflows but occupy the low-throughput region for this training-serving workload. The format is aimed at immutable training corpora; mutable databases and field-projection analytics remain different workloads.

Table~\ref{tab:headline} gives a compact numerical view of the representative 64-atom NVMe slice.

\begin{table}[t]
\caption{Representative 64-Atom NVMe Results}
\label{tab:headline}
\centering
\scriptsize
\begin{tabular}{lrrrr}
\toprule
Backend & Seq. read & Rand. read & Write & Size \\
 & mol/s & mol/s & mol/s & ratio \\
\midrule
Atompack & 646k & 446k & 105k & 1.00 \\
HDF5 SOA & 470k & 18.6k & 91.4k & 0.96 \\
LMDB Packed & 194k & 159k & 37.6k & 2.34 \\
LMDB Pickle & 125k & 117k & 25.0k & 2.35 \\
ASE SQLite & 1.79k & 1.80k & 1.42k & 3.05 \\
ASE LMDB & 4.64k & 4.62k & 0.92k & 4.69 \\
\bottomrule
\end{tabular}
\end{table}

\section{Discussion and Limitations}

The results show that Atompack is most useful when the storage layer has become part of the training bottleneck and when the scientific artifact is naturally an immutable corpus. This is increasingly common in atomistic ML. Large potential-energy-surface datasets are expensive to generate, expensive to move, and some are used widely in benchmarks and training of foundation models. A format optimized for repeated shuffled complete-record access makes it easier to treat these corpora as shared infrastructure rather than as private preprocessing outputs.

Atompack is designed to lower the cost of publishing large-scale atomistic datasets for training. Today, reproducing a model often requires recovering a project-specific conversion script, local database layout, and dataloader assumptions. A common indexed artifact can make a different workflow possible: publish the upstream scientific dataset, publish the exact trainable representation, and allow downstream researchers to benchmark models without first rebuilding the storage layer. This is not only a performance concern. It also improves scientific comparability because storage conversion stops being an unreported part of the experimental method.

The same design choices make Atompack inappropriate for other workloads. It is append-oriented rather than mutable; updates and deletes require rewriting a dataset. It indexes whole molecules rather than arbitrary columns, so it is not a replacement for field-projection analytics over large arrays. It does not provide a query language, transactions across semantic records, or the broad inspection tools of a scientific database. When the bottleneck is graph construction, neighbor-list computation, feature generation, or model training, changing the storage layer will not dominate end-to-end performance.

The benchmarks isolate the storage layer, which is the component Atompack is
designed to improve. Read timings measure indexed record access and field
decoding for fixed atomistic records. Downstream steps such as graph
construction, neighbor-list generation, and model feature construction are not
included; their cost depends on the training code and model architecture. This
scope lets the benchmark compare storage formats directly before model-specific
processing dominates the measurement.

\section{Reproducibility and Availability}

Atompack is available as open-source software at \url{https://github.com/LeMaterial/atompack}. The repository includes the storage implementation, documentation and the benchmark scripts used for evaluation. Public datasets packaged in the Atompack format are distributed through Hugging Face at \url{https://huggingface.co/datasets/LeMaterial/Atompack}.

\section{Conclusion}

Atompack is a focused storage and distribution layer for atomistic ML training datasets. Its central design choice is to store the same unit that training consumes: a complete molecule with its targets. In the maintained benchmark suite, this design gives high shuffled-read and write throughput while keeping artifact size close to compact HDF5 storage. Atompack turns the trainable representation itself into a reusable artifact for local disks, shared filesystems, and public repositories.

\section*{Acknowledgements}

The authors used OpenAI Codex to assist with drafting, editing, and figure-generation code. All benchmark results, implementation details, and conclusions are the authors' own. This work was performed using HPC resources from GENCI-IDRIS and GENCI-CINES under the allocations 2025-A0181016212, 2025-AD011016353, and 2026-A0201016212 on the Jean Zay and Adastra supercomputers.

\bibliographystyle{plainnat}
\bibliography{references}

\end{document}